\documentclass[10pt,twocolumn,letterpaper]{article}

\usepackage[pagenumbers]{iccv} % To force page numbers, e.g. for an arXiv version

\definecolor{iccvblue}{rgb}{0.21,0.49,0.74}
\usepackage[pagebackref,breaklinks,colorlinks,allcolors=iccvblue]{hyperref}
\def\paperID{7} % *** Enter the Paper ID here
\def\confName{ICCV}
\def\confYear{2025}

\title{Beyond Appearance: Geometric Cues for Robust Video Instance Segmentation}

\author{Quanzhu Niu$^{*}$\quad
    Yikang Zhou$^{*}$\quad
    Shihao Chen\quad
    Tao Zhang\quad
    Shunping Ji$^{\dag}$ \\
    {Wuhan University, \ China}\\
    {\tt\normalsize \{quanzhu\_niu, zhouyik, jishunping\}@whu.edu.cn}\\
    {\normalsize\tt\href{https://github.com/QuanzhuNiu/DVIS_Depth}{https://github.com/QuanzhuNiu/DVIS\_Depth}}
}

\usepackage{hyperref}
\usepackage{multirow}
\usepackage{array}
\usepackage{booktabs}
\usepackage{float}
\begin{document}
% \twocolumn[{%
% \renewcommand\twocolumn[1][]{#1}%
% \maketitle
% \begin{center}
%   \centering
%   \begin{subfigure}{0.615\linewidth}
%     \includegraphics[width=1\linewidth]{figs/demos.png}
%   \end{subfigure}
%   \hfill
%   \begin{subfigure}{0.38\linewidth}
%     \includegraphics[width=1\linewidth]{figs/rd3.png}
%   \end{subfigure}
%   \captionof{figure}{Our geometric integration methods enhance robustness in video instance segmentation. (left) Visual comparison of the influence of the proposed EDC method in DVIS-DAQ~\cite{daq}. In the upper part, the baseline DVIS-DAQ method suffers from tracking failures. The person (ID5) on the motorcycle (ID6) is missed in the middle frames and incorrectly associated with the person (ID7) on another motorcycle (ID1). In the lower part, our EDC implementation resolves identity ambiguities and maintains coherent tracking. (right) Quantitative results on OVIS~\cite{OVIS} benchmark demonstrate that the integration of our EDC method universally surpasses the baselines and establishes new state-of-the-art performance. † denotes with the offline refiner proposed by~\cite{DVIS}.}
%   \label{fig:1}
%   \end{center}%
% }]

\twocolumn[{%
\maketitle
\begin{figure}[H]
\hsize=\textwidth
\centering
\vspace{-6mm}
  \begin{subfigure}{1.285\linewidth}
    \includegraphics[width=1\linewidth]{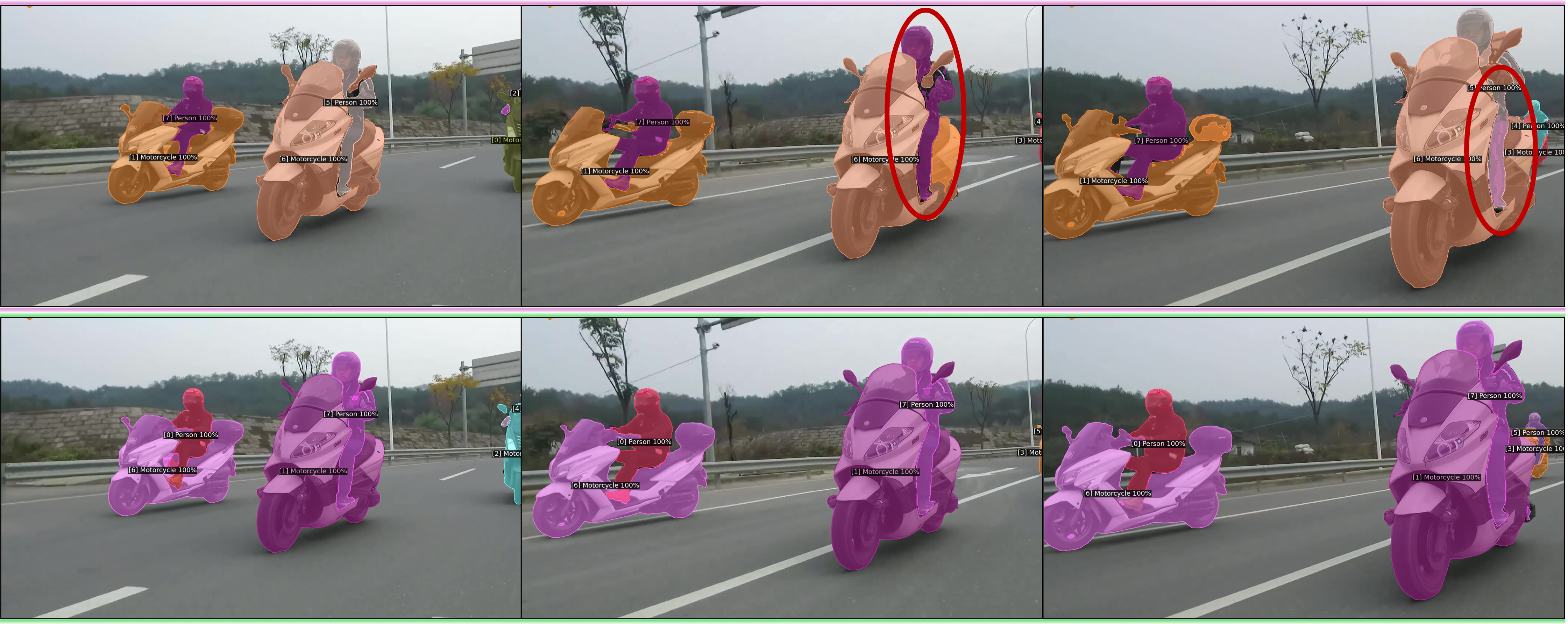}
  \end{subfigure}
  \hfill
  \begin{subfigure}{0.80\linewidth}
    \includegraphics[width=1\linewidth]{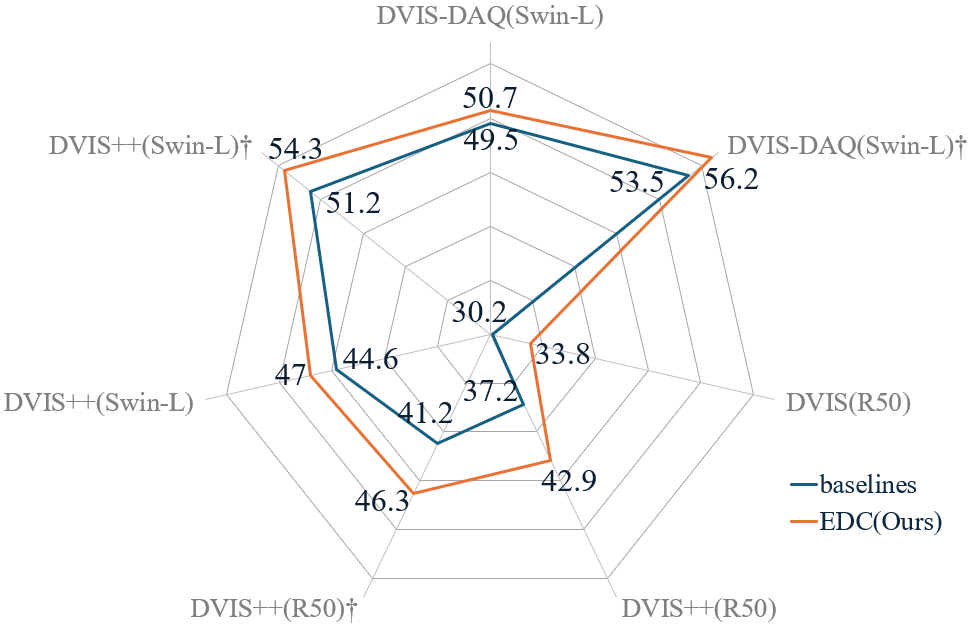}
  \end{subfigure}
  \caption{Our geometric integration methods enhance robustness in video instance segmentation. (left) Visual comparison of the influence of the proposed EDC method in DVIS-DAQ~\cite{daq}. In the upper part, the baseline DVIS-DAQ method suffers from tracking failures. The person (ID5) on the motorcycle (ID6) is missed in the middle frames and incorrectly associated with the person (ID7) on another motorcycle (ID1). In the lower part, our EDC implementation resolves identity ambiguities and maintains coherent tracking. (right) Quantitative results on OVIS~\cite{OVIS} benchmark demonstrate that the integration of our EDC method universally surpasses the baselines and establishes new state-of-the-art performance. † denotes with the offline refiner proposed by~\cite{DVIS}.}
  \label{fig:1}
\end{figure}
\vspace{5mm}
}]

\begin{abstract}
\renewcommand{\thefootnote}{}
\footnote{$^{*}$Equal contribution.}
\footnote{$^{\dag}$Corresponding author.}
Video Instance Segmentation (VIS) fundamentally struggles with pervasive challenges including object occlusions, motion blur, and appearance variations during temporal association. To overcome these limitations, this work introduces geometric awareness to enhance VIS robustness by strategically leveraging monocular depth estimation. We systematically investigate three distinct integration paradigms. Expanding Depth Channel (EDC) method concatenates the depth map as input channel to segmentation networks; Sharing ViT (SV) designs a uniform ViT backbone, shared between depth estimation and segmentation branches; Depth Supervision (DS) makes use of depth prediction as an auxiliary training guide for feature learning. Though DS exhibits limited effectiveness, benchmark evaluations demonstrate that EDC and SV significantly enhance the robustness of VIS. When with Swin-L backbone, our EDC method gets 56.2 AP, which sets a new state-of-the-art result on OVIS benchmark. This work conclusively establishes depth cues as critical enablers for robust video understanding.
\end{abstract}
    
\section{Introduction}
\label{sec:intro}

Video Instance Segmentation (VIS) constitutes a pivotal advancement in computer vision, extending the capabilities of image-level instance segmentation to dynamic video domains~\cite{VIS,OVIS,AwesomeVS}. This task requires simultaneous segmentation, tracking, and recognition of all objects throughout video sequences, producing frame-wise outputs that include spatial positions, semantic categories, and consistent instance identities. Such capabilities underpin critical applications ranging from autonomous navigation systems to intelligent video analytics platforms~\cite{Traj,tf}.

The inherent complexity of VIS arises primarily from temporal challenges: persistent object occlusions, motion blur, and significant appearance variations.  As exemplified in~\cref{fig:1} (left-top), state-of-the-art appearance-driven method~\cite{daq} exhibit catastrophic failure modes under these conditions. In the middle frames of the video, a person is not tracked correctly and is assigned to the ID of another person. Such limitations stem from the fundamental inadequacy of RGB-only features. In some ambiguous situations, however, this dilemma can be alleviated by incorporating beneficial geometrical priors.

As evidenced by previous research~\cite{vip_deeplab,polyphonicformer}, geometric relationships, particularly information about the depth of the scene, provide essential disambiguation cues for these challenges. Depth coherence offers intrinsic advantages, including occlusion reasoning through z-axis ordering
and motion trajectory stabilization via 3D spatial constraints, which effectively resolve ambiguities in crowded scenes under complex motion dynamics. This suggests that integrating depth features could substantially enhance instance embedding learning and cross-frame matching.

Building upon technological advancement~\cite{VIT,DPT,DAv1, DAv2,yang2024depthanyvideo}, the maturation of monocular depth estimation techniques offers new opportunities to strengthen VIS frameworks beyond mere appearance cues and without requiring specialized sensors. Contemporary state-of-the-art depth estimation models~\cite{DAv2,yang2024depthanyvideo} demonstrate unprecedented accuracy in recovering 3D geometry from 2D videos, providing a viable pathway to inject geometric awareness into existing VIS frameworks.   

To enhance robustness beyond appearance cues in video instance segmentation, this work systematically investigates using geometric integration for VIS through three paradigms. 

We propose the Expanding Depth Channel (EDC) method. It integrates depth maps as additional input channels to networks. Surprisingly, this naive fusion strategy achieves significant performance gains. On both R50~\cite{Res} and Swin-L~\cite{Swin} backbones, our EDC method delivers state-of-the-art results.  

We propose the Sharing ViT (SV) method. We use a backbone shared between depth estimation and segmentation streams. This method maximizes feature reuse and yields a superior result than baselines.

We propose the Depth Supervision (DS) method. We try to employ depth prediction as auxiliary training supervision. Though quantitative gains are currently marginal, this method provides a promising direction of utilizing depth in VIS supervision.

As quantified in~\cref{fig:1} (right) and visualized in~\cref{fig:1} (left-bottom), our geometric integration paradigms effectively address the core limitations of appearance-only VIS methods and establishes new performance frontiers on benchmark metrics.

To summarize, our key contributions are as follows:
\begin{itemize}
\item{\textbf{Depth-Aware Architecture Designs.} We propose three approaches to integrate monocular depth predictions with leading VIS frameworks, establishing new frontiers for robust video understanding.}
\item{\textbf{State-of-the-art Performance.} The experiments demonstrate the leading performance of our depth-enhanced frameworks, which establish new state-of-the-art results on prominent VIS benchmarks~\cite{VIS,OVIS}, outperforming existing geometry-agnostic baselines~\cite{DVIS,daq,plus}.}
\end{itemize}

\section{Related Work}
\begin{figure*}
  \centering
  \includegraphics[width=0.815\linewidth]{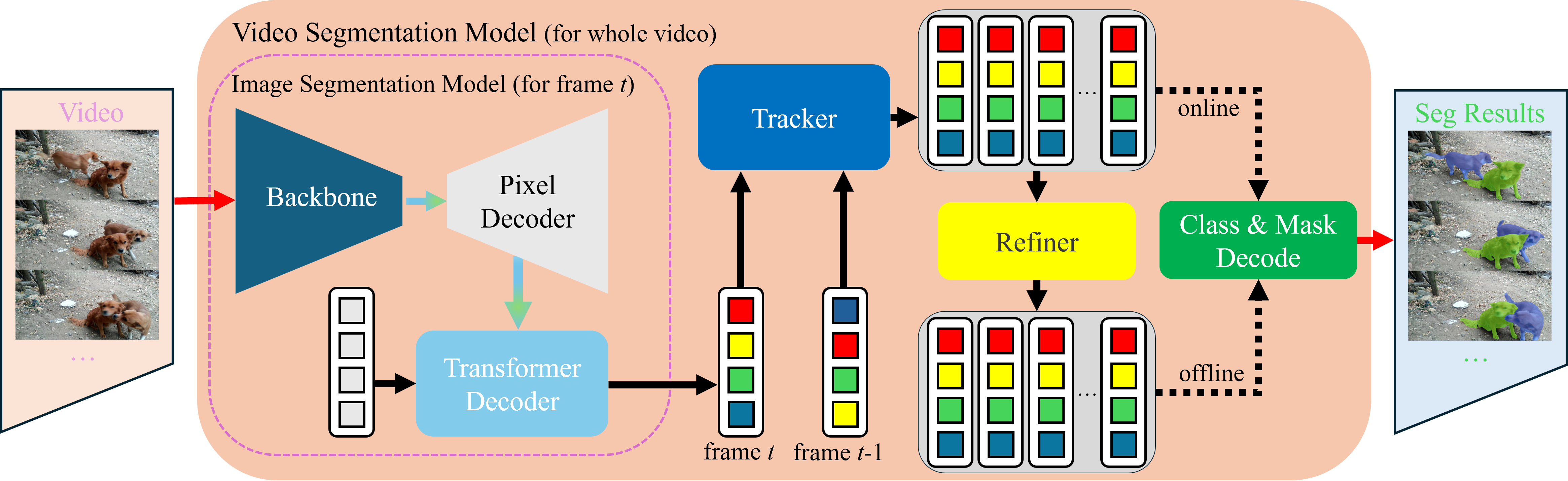}
   \caption{The decoupled framework of video segmentation model established in~\cite{DVIS}, comprising a segmenter, a tracker and a refiner. In online mode, we decode the queries from the tracker. In offline mode, output queries from the refiner are decoded for higher accuracy. }
   \label{fig:pr}
\end{figure*}

\subsection{Video Segmentation}

\textbf{Overview.} Video Segmentation extends image segmentation to video domains, aiming to partition video frames into semantically coherent segments or object groups~\cite{AwesomeVS}. While preserving categorization and pixel-level segmentation capabilities, video segmentation introduces unique challenges, including temporal consistency and cross-frame object association. Following image segmentation, video segmentation tasks include video sementic segmentation (VSS), video instance segmentation (VIS) and video panoptic segmentation (VPS). Unlike three tasks mentioned above, video object segmentation (VOS) aims to track a predefined object across frames~\cite{DAVIS,MOSE} and reference video object segmentation (RVOS) leverages language expressions to identify the target object~\cite{RVOS}. The latter is exemplified by~\cite{MeViS} that utilizes motion expressions as referring cues. Recognizing inherent synergies among these tasks, emerging works~\cite{plus,vknet,OMGSeg} pursue unified frameworks, with~\cite{OMGSeg} demonstrating a versatile architecture capable of handling both image and video segmentation even under open-vocabulary settings.

\noindent
\textbf{Video Instance Segmentation.} The concept of video instance segmentation (VIS) is established by~\cite{VIS}, which first formalizes the task as joint segmentation, tracking, and recognition across video sequences. Subsequent approaches are bifurcated into online and offline paradigms: Online methods~\cite{VIS, MinVIS, IDOL, CTVIS, tcovis} sequentially process frames along the temporal dimension for real-time inference, while offline approaches~\cite{StemSeg, SeqF, VITA, GenVIS} leverage the full-sequence temporal context to enhance precision. With the development of transformers~\cite{Trans} in vision field~\cite{VIT, DETR, DeDETR, MF, M2F}, mainstream VIS methods~\cite{M2Fv, ETE, MinVIS, IDOL, CTVIS, SeqF, VITA, GenVIS, GRAtt, tcovis, DVIS, plus, daq, cavis} turn to query-based design. Recent breakthroughs focus on decoupled designs \cite{DVIS, plus, daq}, which delegate segmentation to query-based image models~\cite{M2F} and manage tracking via memory-augmented transformers. They could support both online and offline modes. Specifically, a tracker enables real-time operation in online mode, while an optional offline refiner enhances results through temporal consistency. \cite{plus} utilizing progressive denoising with contrastive learning for query propagation. \cite{daq} focuses on solving the problem of newly emerging and disappearing objects in videos using a dynamic anchor query mechanism. \cite{cavis} enhances instance matching through context-aware instance features. Critically, under severe occlusions, appearance-driven methods fail to distinguish instances with homogeneous textures, leading to degraded matching reliability due to insufficient geometric constraints.

\subsection{Monocular Depth Estimation}

Monocular depth estimation (MDE) aims to reconstruct reliable 3D geometry from 2D images or videos without specialized sensors, with recent advances predominantly driven by deep learning paradigms. Early CNN-based methods pioneered supervised learning frameworks, exemplified by~\cite{RMSE} which introduced scale-invariant loss functions to optimize depth regression accuracy, while~\cite{DORN} reformulated depth prediction as an ordinal regression problem. Subsequent architectural innovation~\cite{DPT} uses Vision Transformer (ViT)~\cite{VIT} to replace convolutional networks in depth prediction tasks. A significant research shift has emerged toward zero-shot relative depth estimation, where transformative progress stems from large-scale pre-training strategies~\cite{MiDaS,m3.1,M3D,M3Dv2,DAv1,DAv2}. \cite{DAv2} significantly expanded training diversity through synthetic image augmentation. Most recently, \cite{yang2024depthanyvideo} establishes a novel video monocular depth estimation framework that leverages a video generation model~\cite{blattmann2023stablevideodiffusionscaling}.

\subsection{Depth and Segmentation Integration}
\begin{figure*}
  \centering
    \includegraphics[width=1\linewidth]{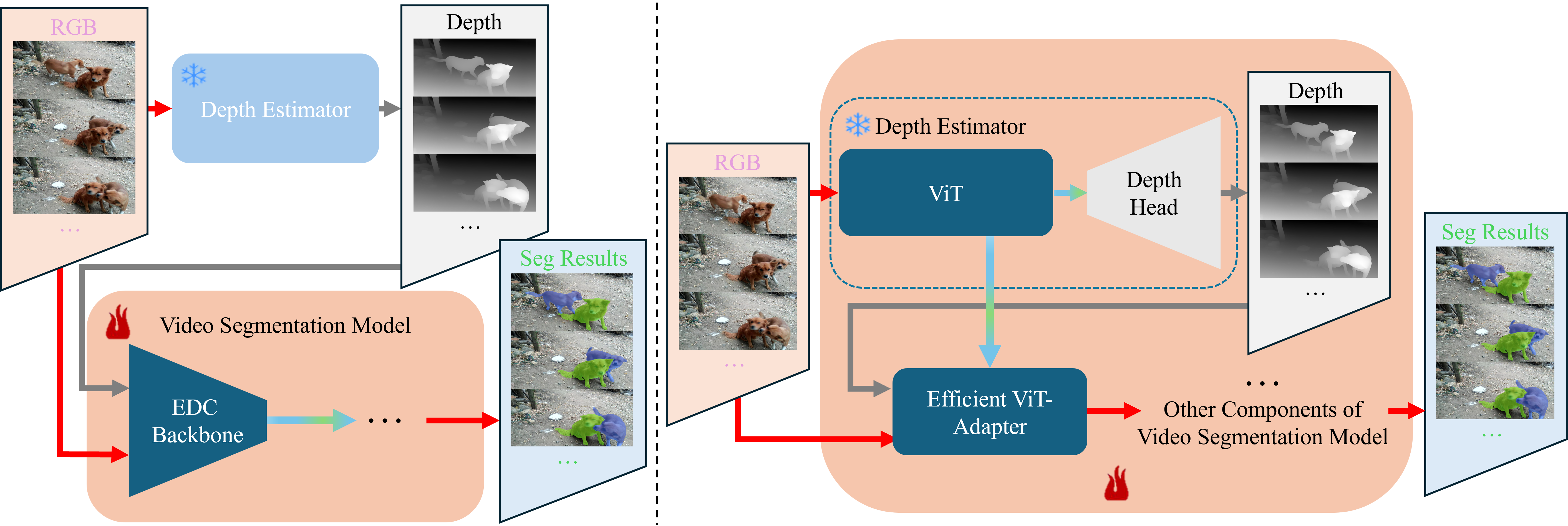}
  \caption{(left) The architecture of Expanding Depth Channel (EDC) method. The 4-channel EDC backbone extracts spatiotemporal features from RGB-D video inputs. (right) The architecture of Sharing ViT (SV) method. Both segmentation stream and depth stream share a frozen pre-trained ViT~\cite{dinov2, VIT}. }
  \label{fig:4csv}
\end{figure*}

Prior research has explored various paradigms for joint depth estimation and segmentation. Early works~\cite{Chen_2019_CVPR, ramirez2018, Saeedan_2021_WACV, sdc, sg} established multi-task learning frameworks or utilized segmentation to enhance monocular depth prediction. Subsequent approaches advanced this synergy through architectural innovations. \cite{vip_deeplab} extended panoptic segmentation architectures~\cite{Cheng_2020_CVPR} with dedicated depth prediction heads, while~\cite{polyphonicformer} introduced a unified transformer decoder that processes both tasks through shared object queries, enabling bidirectional feature refinement. What's more, some works~\cite{zhang2023cmx,zhang2023delivering,Wan_2025_WACV} explore fusing RGB with depth and other modals such as thermal for segmentation. Concurrently, multi-modal fusion techniques~\cite{zhang2023cmx, zhang2023delivering, Wan_2025_WACV} have demonstrated performance gains by integrating RGB with complementary modalities such as depth or thermal for robust segmentation. Collectively, these efforts establish geometric awareness as a critical enhancement for segmentation systems, though they typically retain explicit dependency on ground-truth depth data.
\section{Method}

\subsection{Preliminary}

Contemporary video segmentation predominantly adopts query-based architectures, extending image-level frameworks~\cite{MF, M2F} through temporal association modules for cross-frame instance matching~\cite{AwesomeVS}. As illustrated in~\cref{fig:pr}, our implementation follows the decoupled paradigm established in DVIS~\cite{DVIS}, consisting of: (1) a segmenter built upon the image segmentation model; (2) a tracker to propagate object associations across adjacent frames; and (3) for offline processing, a refiner applied to the tracker outputs to optimize temporal consistency throughout video sequences. The instance queries can be decoded into masks and class labels, which are the video instance segmentation results.

\subsection{Expanding Depth Channel}
\label{sec:3edc}

The most straightforward approach for leveraging depth information is to feed the depth map as a new channel into the network. We propose an Expanding Depth Channel (EDC) method that synergistically integrates RGB appearance cues with geometric depth priors. Illustrated in \cref{fig:4csv} (left), the proposed method employs a channel expansion strategy. When ground-truth depth maps are unavailable, we leverage monocular depth estimation model~\cite{DAv2} to derive additional depth features.

Specifically, we implement state-of-the-art Depth Anything V2 model~\cite{DAv2} as our Depth Estimator. The Depth Estimator generates relative depth values in pixels. At one specific frame, the estimated depth map $D\in \mathbb{R} ^{H\times W}$ is concatenated with the original 3-channel RGB image $I\in \mathbb{R} ^{3\times H\times W}$ along the channel dimension, constructing a 4-channel RGBD frame. This constructs a spatiotemporal RGBD video $\hat{V}\in \mathbb{R} ^{T\times 4\times H\times W}$.

To accommodate this multi-modal input, we modify the backbone of the baseline video segmentation model through channel dimension expansion. As shown in \cref{fig:4csv} (left), the first layer of the backbone turns to get the input of 4-channel images while preserving pre-trained parameters for original RGB streams. This design enables seamless integration of geometric constraints without compromising the model's ability to learn appearance-based features. During both training and inference phases, we send 4-channel videos into the video segmentation model to get the final segmentation results.
\begin{figure*}
  \centering
  \includegraphics[width=0.75\linewidth]{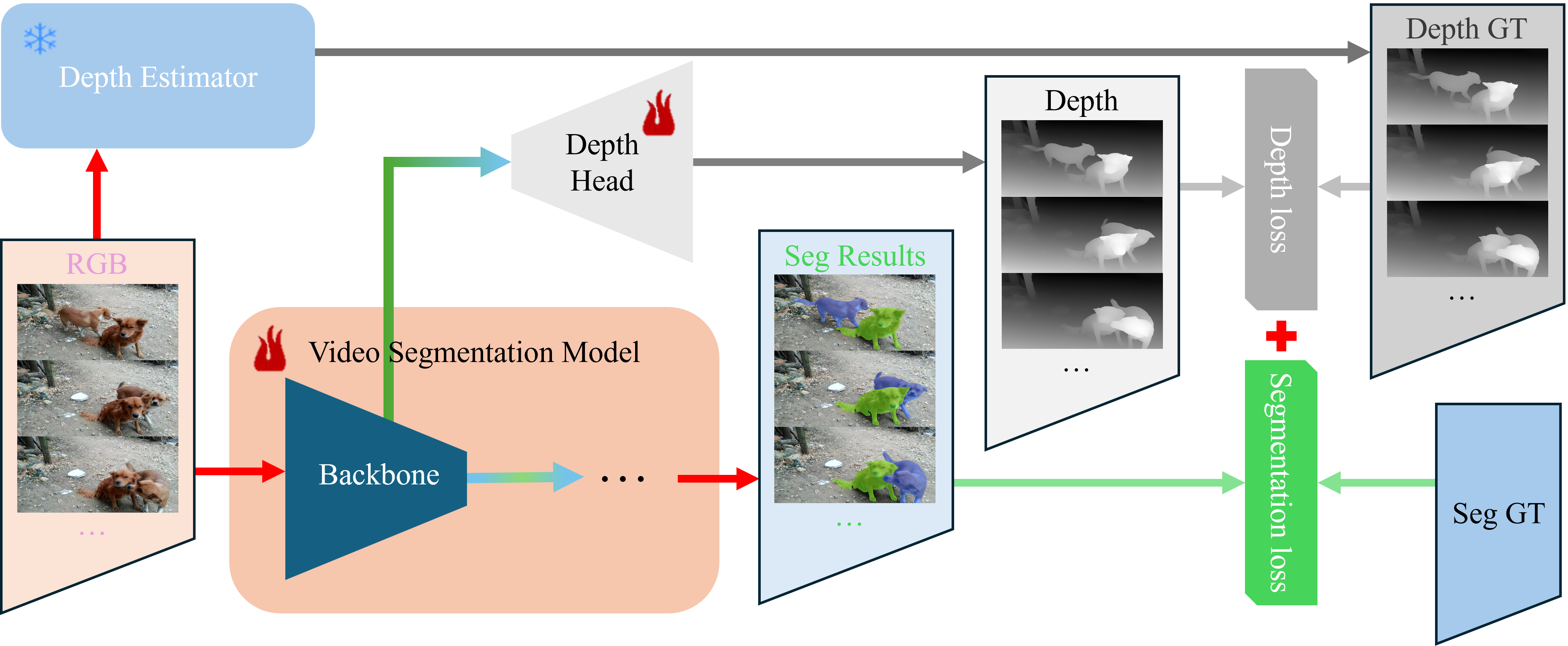}
   \caption{The architecture of Depth Supervision (DS) method. Both depth estimation and segmentation losses are jointly optimized during multi-task learning. }
   \label{fig:ds}
\end{figure*}

\subsection{Sharing ViT}

While current state-of-the-art video segmentation methods~\cite{plus, daq, cavis} attain optimal performance by leveraging DINOv2 ~\cite{dinov2} pre-trained Vision Transformer (ViT)~\cite{VIT} with ViT-Adapter~\cite{adapter}, computational constraints often necessitate maintaining the ViT backbone in a frozen state. Meanwhile, depth estimation models inherently require backbone features for RGB-to-depth mapping. Thus, we propose a method that strategically couples the monocular depth estimation model with the segmentation model. Our design enables parameter-sharing of the powerful ViT backbone between both tasks, thereby enhancing segmentation capability while preserving depth estimation accuracy.

The Sharing ViT (SV) architecture, as illustrated in \cref{fig:4csv} (right), employs the frozen Depth Anything V2~\cite{DAv2} model. The pre-trained ViT simultaneously serves two purposes: (1) providing features for the depth prediction head through its original pathway, and (2) delivering multi-scale hierarchical features to the segmentation branch via a efficient version ViT-Adapter~\cite{adapter} module (with injector components removed). Building upon the EDC method detailed in~\cref{sec:3edc}, we implement a streamlined integration approach where the depth map generated by the depth prediction head is concatenated as an additional input channel to the spatial prior module within the ViT-Adapter~\cite{adapter}.This operation constructs spatio-temporal RGBD video inputs $\hat{V}\in \mathbb{R} ^{T\times 4\times H\times W}$, where the fourth channel encodes geometric priors.

This design achieves dual-functional optimization. First, geometric awareness is preserved by leveraging the frozen ViT’s~\cite{dinov2,VIT} capacity for depth-aware feature extraction. Then, it maintains the efficiency of the fusion by enabling cross-modal interaction between appearance (RGB) and geometry (D) through the ViT-Adapter~\cite{adapter}.

\subsection{Depth Supervision}

To eliminate reliance on external depth estimation models during inference, we investigate Depth Supervision (DS) as a training strategy to enhance geometric awareness in video segmentation. Despite conducting extensive experiments employing depth prediction results as auxiliary supervision in diverse architectural configurations, most implementations proved insufficient to enhance segmentation performance.

As shown in~\cref{fig:ds}, our framework processes multi-scale backbone features through two parallel objectives: (1) primary segmentation supervised by ground truth labels, and (2) auxiliary depth prediction supervised by pseudo labels from~\cite{DAv2}. The depth head follows the architecture in~\cite{DPT}. Crucially, both branches share identical backbone features but employ task-specific heads, enabling joint optimization without cross-task interference. We show the results of this strategy in~\cref{sec:4ds}.

We further explore an architectural extension where the transformer decoder in our image segmentation model jointly decodes object queries for both segmentation masks and depth prediction. This multi-task decoding framework leverages pseudo-depth supervision from~\cite{DAv2} to regularize instance-level depth estimation. Despite successful architectural integration, empirical evaluation reveals poor performance, suggesting inherent challenges in using depth supervision in the segmentation task.

\subsection{Implementation Details}
\label{sec:3id}

Our three depth-aware methodologies are built upon the DVIS series~\cite{DVIS, plus, daq} video segmentation frameworks. The four-stage training protocol follows: (1) Training the image segmentation model~\cite{M2F}; (2) Initializing the video segmenter using the pre-trained image model followed by fine-tuning on video instance segmentation datasets~\cite{MinVIS, CTVIS}; (3) Freezing the segmenter while optimizing the tracker; (4) Training the refiner with both segmenter and tracker frozen.

We adhere to baselines' training and inference configurations. The AdamW optimizer~\cite{adamw} is employed with an initial learning rate of 1.0e-4 and a weight decay of 5.0e-2. For the training of image segmentation model~\cite{M2F} on COCO~\cite{coco} dataset, ResNet-50 (R50)~\cite{Res} and ViT-L~\cite{VIT} backbones undergo 50 epochs, while Swin-L~\cite{Swin} requires 100 epochs. Video segmentation training utilizes COCO pseudo-videos~\cite{SeqF} for joint training on OVIS~\cite{OVIS}, YouTube-VIS 2019 and YouTube-VIS 2021~\cite{VIS} datasets. The training iteration settings are: (1) DVIS~\cite{DVIS}: 20k for OVIS and 40k for YouTube-VIS; (2) DVIS++~\cite{plus}: 40k; (3) DVIS-DAQ~\cite{daq}: 160k.

\section{Experiments}

\subsection{Datasets and Metrics}

Our experiments are conducted on three established video instance segmentation benchmarks: OVIS~\cite{OVIS}, YouTube-VIS 2019 and YouTube-VIS 2021~\cite{VIS}. The OVIS dataset is characterized by challenging real-world scenarios including severe occlusions, rapid object motion, and complex motion trajectories. The YouTube-VIS 2019 and 2021 datasets exhibit constrained temporal scope and simplified scenario compositions, providing standardized evaluation for short-term segmentation consistency. 

We adopt the standard Average Precision (AP) and Average Recall (AR) metrics as defined in~\cite{VIS} for video instance segmentation task.  
\begin{table*}
  \centering
  \begin{tabular}{ll|ccc|ccc|ccc}
\toprule
\multicolumn{2}{c|}{\multirow{2}{*}{Method}} &
  \multicolumn{3}{c|}{OVIS} &
  \multicolumn{3}{c|}{YouTube-VIS 2019} &
  \multicolumn{3}{c}{YouTube-VIS 2021} \\
\multicolumn{2}{c|}{} &
  \multicolumn{1}{c}{AP} &
  \multicolumn{1}{c}{AP$_{75}$} &
  \multicolumn{1}{c|}{AR$_{10}$} &
  \multicolumn{1}{c}{AP} &
  \multicolumn{1}{c}{AP$_{75}$} &
  \multicolumn{1}{c|}{AR$_{10}$} &
  \multicolumn{1}{c}{AP} &
  \multicolumn{1}{c}{AP$_{75}$} &
  \multicolumn{1}{c}{AR$_{10}$} \\ \midrule
\multirow{2}{*}{DVIS~\cite{DVIS}}      & baseline    &30.2&30.5&37.3& 51.2 &57.1&59.3& 46.4 &49.6&53.5\\
                           & EDC      & \textbf{33.8}(+3.6)&\textbf{32.1}&\textbf{40.6}& \textbf{55.9}(+4.7) &\textbf{63.8}  &\textbf{64.8}  & \textbf{50.9}(+4.5) &\textbf{57.5}  &\textbf{58.3}  \\ \midrule
\multirow{3}{*}{DVIS++~\cite{plus}}    & baseline    & 37.2 &37.3&\underline{42.9}& 55.5 &60.1&62.6& 50.0 &54.5&58.4\\
                           & EDC       & \textbf{42.9}(+5.7) &\textbf{42.3}&\textbf{48.7}& \textbf{58.1}(+2.6) &\textbf{65.9}  &\textbf{65.9}  & \textbf{53.0}(+3.0) &\textbf{60.1}  &\textbf{60.7}  \\ 
                           & DS& \underline{37.2}&\underline{38.0}&42.8& - &-  &-  & - &-  &-  
                             \\ \midrule
\multirow{3}{*}{DVIS++†~\cite{plus}}   & baseline    & \underline{41.2} &40.9&\underline{47.3}& 56.7 &62.0&64.7& 52.0 &57.8&59.6\\
                           & EDC       & \textbf{46.3}(+5.1) &\textbf{46.9}&\textbf{52.1}& \textbf{59.1}(+2.4) &\textbf{66.7}  &\textbf{67.3}  & \textbf{54.1}(+2.1) &\textbf{61.2}  &\textbf{61.8}  \\ 
                           & DS&40.1&\underline{41.0}&46.0& - &-  &-  & - &-  &-  \\ \bottomrule
\end{tabular}
  \caption{Results on the valid set of OVIS and YouTube-VIS 2019 \& 2021 with R50~\cite{Res} backbone. † denotes with the offline refiner proposed by~\cite{DVIS}. }
  \label{tab:results1}
\end{table*}

\begin{table*}
  \centering
  \begin{tabular}{ll|l|ccc}
\toprule
\multicolumn{2}{c|}{\multirow{2}{*}{Method}} &
  \multicolumn{1}{c|}{\multirow{2}{*}{Backbone}} &
  \multicolumn{3}{c}{OVIS}  \\
\multicolumn{2}{c|}{} &
  \multicolumn{1}{c|}{} &
  \multicolumn{1}{c}{AP} &
  \multicolumn{1}{c}{AP$_{75}$} &
  \multicolumn{1}{c}{AR$_{10}$}  \\ \midrule
\multirow{4}{*}{DVIS++~\cite{plus}}    & baseline & Swin-L &44.6&46.4&49.7\\
                           & EDC      & Swin-L &\textbf{47.0}(+2.4)&\textbf{49.6}&\textbf{52.0}\\ \specialrule{0em}{1pt}{1pt} \cline{2-6}
                           \specialrule{0em}{1pt}{1pt}
                           &baseline & ViT-L  &49.6&\textbf{55.0}&54.6 \\ 
                           &SV & ViT-L  &\textbf{50.1}(+0.5)&52.4 &\textbf{54.9} \\ \midrule
\multirow{4}{*}{DVIS++†~\cite{plus}}  & baseline & Swin-L &51.2&53.8&55.9\\
                           & EDC      & Swin-L &\textbf{54.3}(+3.1)&\textbf{57.7}&\textbf{59.3}   \\ \specialrule{0em}{1pt}{1pt} \cline{2-6}
                           \specialrule{0em}{1pt}{1pt} 
                           &baseline& ViT-L  &53.4&58.5&58.7 \\
                           &SV & ViT-L  &\textbf{55.8}(+2.4)&\textbf{59.1}&\textbf{60.5}\\ \midrule  
\multirow{2}{*}{DVIS-DAQ~\cite{daq}}  & baseline & Swin-L & 49.5 & 51.7 & 54.9  \\
                           & EDC      & Swin-L & \textbf{50.7}(+1.2) & \textbf{54.3}& \textbf{56.4} \\ \midrule
\multirow{2}{*}{DVIS-DAQ†~\cite{daq}}  & baseline & Swin-L & 53.5     &  58.0    & 59.0 \\
                           & EDC      & Swin-L & \textbf{56.2}(+2.7) & \textbf{61.9} & \textbf{60.6}
                           \\ \bottomrule
\end{tabular}
  \caption{Results on the valid set of OVIS with Swin-L~\cite{Swin} and ViT-L~\cite{VIT} backbones. † denotes with the offline refiner proposed by~\cite{DVIS}. The ViT-L~\cite{VIT} is pre-trained by DINOv2~\cite{dinov2} and uses ViT-Adapter~\cite{adapter} to obtain multiscale features.}
  \label{tab:results2}
\end{table*}

\begin{table*}[]
    \centering
   \begin{tabular}{ll|ccc}
\toprule
\multirow{2}{*}{Method}   & \multirow{2}{*}{Depth Estimate Model} & \multicolumn{3}{c}{OVIS} \\ & &AP &AP$_{75}$&AR$_{10}$ \\ \midrule
\multirow{2}{*}{DVIS++-EDC}    & Depth-Anything-V2-Large (335M)  & 42.9 &42.3&48.7\\
                           &  Depth-Anything-V2-Small (25M)      & 39.9(-3.0) &39.5&46.0  \\ \midrule
\multirow{2}{*}{DVIS++-EDC†}    & Depth-Anything-V2-Large (335M)  & 46.3 &46.9&52.1\\
                           & Depth-Anything-V2-Small (25M)      & 43.1(-3.2) &42.3&49.6  \\   \bottomrule
\end{tabular}
    \caption{Ablation study on effect of the quality of the depth map by using different depth estimate models~\cite{DAv2}. We evaluate on R50~\cite{Res} backbone. † denotes with the offline refiner proposed by~\cite{DVIS}.}
    \label{tab:ab1}
\end{table*}

\begin{table}[]
    \centering
   \begin{tabular}{l|ccc}
\toprule
\multirow{2}{*}{Add EDC Stage} & \multicolumn{3}{c}{OVIS} \\ &AP &AP$_{75}$&AR$_{10}$ \\ \midrule
 Image Seg Model  & 39.0 &39.0&42.8\\
                        Segmenter      & 33.2(-5.8) &33.3&36.9  \\ \midrule  
                     w/o EDC     & 35.5(-3.5) &36.1&38.9  \\  \bottomrule
\end{tabular}
    \caption{Ablation study on effect of EDC initialization stages using Swin-L~\cite{Swin} backbone. Performance is evaluated under three configurations: Image Seg Model denotes initializing depth in image segmentation stage; Segmenter denotes initializing depth in video segmenter stage; w/o EDC denotes the baseline without EDC in any stage. }
    \label{tab:ab2}
\end{table}

\subsection{Expanding Depth Channel Results}

\textbf{Performance on OVIS dataset.} Quantitative evaluations in~\cref{tab:results1} and~\cref{tab:results2} demonstrate that integrating depth information through our Expanding Depth Channel (EDC) method yields substantial VIS accuracy improvements across multiple baselines. The most substantial enhancement is observed with the R50 backbone~\cite{Res}, where the EDC method outperforms baseline DVIS++~\cite{plus} by 5.7 AP. Notably, when incorporating EDC within DVIS++~\cite{plus} offline, the Swin-L~\cite{Swin} backbone achieves 54.3 AP, even surpassing the ViT-L~\cite{VIT} counterpart by 0.9 AP. Furthermore, when using Swin-L backbone, our EDC method integrated with DVIS-DAQ~\cite{daq} with offline refiner establishes a new state-of-the-art results of 56.2 AP on OVIS. This result outperforming any existing method with Swin-L backbone, demonstrating our EDC method's synergistic capacity to leverage geometric priors.

\noindent
\textbf{Performance on YouTube-VIS 2019 \& 2021 datasets.}  Quantitative comparisons are shown in~\cref{tab:results1}. On both benchmarks, our EDC method consistently achieves measurable improvements over baselines. When integrating EDC in DVIS~\cite{DVIS}, performance increase by at least 4.5 AP. The results demonstrate that our EDC method achieves systematic performance gains across all metrics, statistically validating its efficacy in improving segmentation consistency under low-complexity conditions.

\subsection{Sharing ViT Results}

The proposed Sharing ViT (SV) method demonstrates significant improvements through depth-aware feature fusion, yielding substantial quantitative gains over baseline models. As illustrated in~\cref{tab:results2}, on the challenging OVIS benchmark, DVIS++~\cite{plus} equipped with SV achieves 53.4 AP, which surpasses the baseline by 3.1. Furthermore, when integrated with an offline refiner~\cite{DVIS}, the SV-enhanced DVIS++ attains 55.8 AP, outperforming the baseline by 2.4 AP. These consistent gains across both online and offline configurations validate SV’s efficacy in leveraging geometric priors to enhance segmentation robustness under dynamic scenarios.

\subsection{Depth Supervision Results}
\label{sec:4ds}

Quantitative comparisons in~\cref{tab:results1} reveal the Depth Supervision (DS) approach (illustrated in~\cref{fig:ds}) achieves performance parity with the baseline model. On OVIS benchmark, DVIS++~\cite{plus} augmented with DS exhibits a marginal but consistent improvement AP$_{75}$, indicating enhanced localization precision for high-confidence predictions. While both depth and segmentation tasks achieved successful convergence, the overall segmentation accuracy shows no measurable improvement compared to the baseline without depth supervision, suggesting the limited effectiveness of current DS approaches.

\subsection{Ablation Studies}

\textbf{Quality of the depth map.} The explicit integration of depth maps into the backbone architecture establishes a measurable interdependence between depth estimation quality and segmentation performance. As shown in~\cref{tab:ab1}, we quantify this dependency by generating depth maps using multi-scale variants of Depth Anything V2~\cite{DAv2} (large and small) and integrating them through our Expanding Depth Channel (EDC) framework. When employing large depth estimate model, the EDC-enhanced DVIS++~\cite{plus} can outperform the small depth estimate model by not less than 3.0 AP. Quantitative analysis confirms a quantifiable relationship where enhanced depth map accuracy consistently improves video instance segmentation metrics. 

\noindent
\textbf{Start training stage for EDC.} As detailed in~\cref{sec:3id}, we adopt a four-stage training protocol. To determine the optimal training strategy for EDC, we compare two approaches: (1) Training the image segmentation model with EDC from initialization; (2) Initializing the segmenter with a standard 3-channel pre-trained image segmentation model and training it with EDC. Quantitative comparisons (see~\cref{tab:ab2}) demonstrate the superiority of early EDC integration. When applying EDC during image segmentation model training, the segmenter achieves 39.0 AP on OVIS, surpassing late-stage EDC adaptation by 5.8 AP.  Notably, introducing EDC only at the segmenter stage degrades performance by 2.3 AP compared to the pure RGB baseline. This confirms the critical importance of fusing depth information at the earliest training stage.

\begin{figure*}[t]
  \centering
  \includegraphics[width=1.018\linewidth]{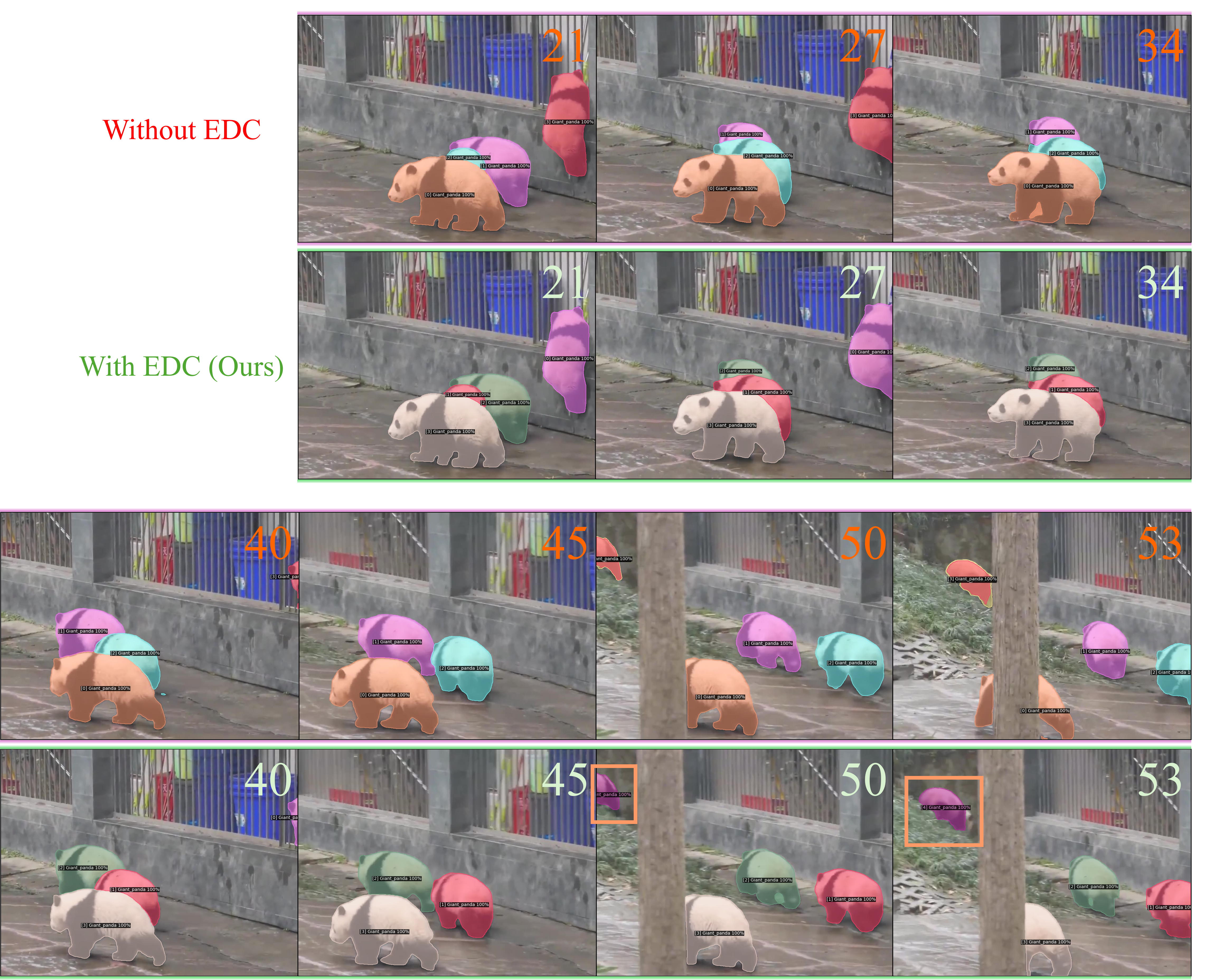}
   \caption{Qualitative comparison of tracking robustness between DVIS-DAQ~\cite{daq} baseline and our EDC method on an OVIS~\cite{OVIS} test video. The frame indices are on the top-right corner of each frame image. (top row) DVIS-DAQ~\cite{daq} erroneously associates a newly appeared panda with a disappeared panda (ID3), violating temporal coherence. (bottom row) Our EDC-enhanced framework correctly distinguishes the two instances by leveraging geometric consistency, as highlighted by orange rectangles. }
   \label{fig:dl}
\end{figure*}
\noindent
\textbf{Qualitative analysis.} As illustrated in~\cref{fig:dl}, we evaluate our EDC method against the previous state-of-the-art approach DVIS-DAQ~\cite{daq} on a representative video from the OVIS~\cite{OVIS} test set. The top row depicts tracking results from DVIS-DAQ~\cite{daq}, while the bottom row demonstrates our EDC-enhanced framework.

In the baseline track (DVIS-DAQ~\cite{daq}), a critical identity discontinuity occurs during camera leftward panning: the panda initially assigned ID3 (denoted as the disappeared panda) disappears after frame No.40. Subsequently, at frame No.50, a new panda emerges from the left viewport (denoted as the newly appeared panda). Crucially, DVIS-DAQ~\cite{daq} erroneously associates this new instance with the original ID3 label from the disappeared panda, violating temporal consistency. In contrast, our EDC method maintains rigorous identity coherence. The disappeared panda retains ID0 until exit, while the newly appeared panda is correctly assigned a distinct ID4, preserving instance-level correctness.

These observations substantiate the fact that geometric-aware modeling significantly mitigates the inherent limitation of appearance-driven VIS methods. While temporal logic dictates that the disappeared panda cannot physically translocate to the new position within the video's spatio-temporal constraints, appearance-centric matching fails to distinguish the two pandas due to their high visual similarity in color and texture. However, depth augmentation in our methods provides decisive geometric disambiguation, revealing that the disappeared panda disappears close to the camera, while the newly appeared panda emerges from a spatially independent faraway region. Thus, our depth-aware methods resolves tracking failures and achieves superior VIS performance.

\section{Conclusion}

This work conclusively demonstrates that geometric cues are indispensable for robust video instance segmentation under challenging conditions, redefining geometric integration paradigms for robust video understanding. Through rigorous experimentation, we demonstrate: Early fusion of depth information via Expanding Depth Channel (EDC) substantially enhances segmentation consistency among all methods, with AP gains of up to 5.7; Cross-modal feature sharing in Sharing ViT (SV) optimally leverages frozen ViT capacities, enabling efficient depth-RGB feature interaction without architectural expansion as EDC; Depth Supervision (DS) fails to translate geometric awareness into segmentation improvements, revealing limitations in auxiliary task alignment. 

A promising direction involves developing enhanced depth supervision mechanisms including geometry-aware contrastive learning to eliminate inference-time dependency on depth information. We hope for further investigation above our researches.

{
    \small
    \bibliographystyle{ieeenat_fullname}
    \bibliography{main}
}

\end{document}